\documentclass{article}

    \usepackage[nonatbib, preprint]{neurips_2021}

\usepackage[utf8]{inputenc} 
\usepackage[T1]{fontenc}    
\usepackage{hyperref}       
\usepackage{url}            
\usepackage{booktabs}       
\usepackage{amsfonts}       
\usepackage{nicefrac}       
\usepackage{microtype}      
\usepackage{xcolor}         
\usepackage{graphicx}
\usepackage{float}
\usepackage{amsmath}
\title{On the role of feedback in visual processing: a predictive coding perspective}

\author{%
  Andrea Alamia\thanks{Equal Contribution} \\
  CerCo, CNRS, 31052 Toulouse, France\\
  \texttt{andrea.alamia@cnrs.fr} \\
  \And
  Milad Mozafari\textsuperscript{*} \\
  CerCo, CNRS, 31052 Toulouse, France\\
  IRIT, CNRS, 31062, Toulouse, France \\
  \texttt{milad.mozafari@cnrs.fr} \\
  \And
  Bhavin Choksi \\
  CerCo, CNRS, 31052 Toulouse, France\\
  \texttt{bhavin.choksi@cnrs.fr} \\
  \And
  Rufin VanRullen \\
  CerCo, CNRS, 31052 Toulouse, France\\
  ANITI, Universit\'e de Toulouse, 31062, France\\
  \texttt{rufin.vanrullen@cnrs.fr} \\
}

\begin{document}

\maketitle

\begin{abstract}
   Brain-inspired machine learning is gaining increasing consideration, particularly in computer vision. Several studies investigated the inclusion of top-down feedback connections in convolutional networks; however, it remains unclear how and when these connections are functionally helpful. Here we address this question in the context of object recognition under noisy conditions. We consider deep convolutional networks (CNNs) as models of feed-forward visual processing and implement Predictive Coding (PC) dynamics through feedback connections (predictive feedback) trained for reconstruction or classification of clean images. To directly assess the computational role of predictive feedback in various experimental situations, we optimize and interpret the hyper-parameters controlling the network's recurrent dynamics. That is, we let the optimization process determine whether top-down connections and predictive coding dynamics are functionally beneficial. Across different model depths and architectures (3-layer CNN, ResNet18, and EfficientNetB0) and against various types of noise (CIFAR100-C), we find that the network increasingly relies on top-down predictions as the noise level increases; in deeper networks, this effect is most prominent at lower layers. In addition, the accuracy of the network implementing PC dynamics significantly increases over time-steps, compared to its equivalent forward network. All in all, our results provide novel insights relevant to Neuroscience by confirming the computational role of feedback connections in sensory systems, and to Machine Learning by revealing how these can improve the robustness of current vision models.
\end{abstract}

\section{Introduction}
\label{intro}

Feed-forward deep convolutional networks (DCNs) reached remarkable accuracy in several visual tasks, including image classification. Interestingly, DCNs share several similarities with biological visual systems. For example, both systems have a hierarchical structure, in which neurons in the higher (lower) levels of the hierarchy have larger (smaller) receptive field sizes and respond to more complex (simpler) stimuli~\cite{hubel1959receptive}. Further, representational~\cite{khaligh2014deep} and functional similarities~\cite{bashivan2019neural} between the feed-forward DCNs and the brain's feed-forward visual pathway have provided novel opportunities to study the brain through the lens of DCNs.

However, contrary to biological visual systems, DCNs blunder significantly when confronted with noisy images and adversarial attacks, revealing an important deficit in robustness~\cite{szegedy2013intriguing,nguyen2014deep,hendrycks2019benchmarking}. One main difference with their biological counterpart consists in the lack of recurrent or feedback connections. It has been shown that the brain relies on feedback pathways for robust object recognition under challenging conditions~\cite{wyatte2014early,kietzmann2019recurrence,kar2019evidence,rajaei2019beyond,kar2021fast}. In recent years, several approaches aimed to introduce feedback connections in deep networks to improve not only biological plausibility but also model robustness, and accuracy~\cite{huang2020neural, kubilius2018cornet, nayebi2018task, yan2019recurrent}. Importantly, feedback connections can be trained either in a supervised fashion to optimize the task objective (e.g., object recognition) or in an unsupervised way to minimize the reconstruction errors (i.e., prediction errors). In the latter case, feedback connections are trained to predict the activity of lower layers, and the network can be described as a hierarchical generative model. More generally, top-down predictions represent prior expectations about lower layers activity, updated based on the incoming sensory evidence over iterations. This interpretation about the role of top-down connections finds its natural place in a prominent framework in Neuroscience, namely Predictive Coding~\cite{huang2011predictive, rao1999predictive}.

The Predictive Coding (PC) paradigm in Neuroscience is endorsed by a large body of neuroscientific experimental evidence~\cite{kilner2007predictive, baldeweg2006repetition, garrido2009mismatch, hohwy2008predictive}. It characterizes perception as an inference process in which sensory information is combined with prior expectations to attain the final percept. Accordingly, PC postulates two fundamental terms: predictions and prediction errors (PEs). Considering the visual system as a hierarchical structure, these two signals interact between subsequent brain regions in an iterative process. Ideally, the interplay between feedback predictions and feed-forward PEs converges over iterations into a state in which predictions fully represent the sensory information and PE falls to zero. Although several models implemented and described this dynamic in different conditions~\cite{friston2009predictive, alamia2019alpha, spratling2010predictive}, the functional role of these two main actors remains largely unexplored.

Here, we address this question by taking a computational perspective and leveraging current state-of-the-art deep neural networks used in visual object recognition. The key insight in our approach consists in letting the network decide for itself (through hyper-parameter optimization) whether top-down connections are functionally beneficial; we then evaluate the outcome across various experimental (noise) conditions. On the one hand, from a Neuroscience point of view, our results supported the hypothesis that feedback plays a crucial role in the cortical processes involved in biological vision. On the other hand, from a machine learning perspective, our simulations demonstrated a more robust class of models based on an established biologically inspired framework.

\section{Methods}
\label{methods}

\subsection{Predictive Coding Dynamics}
Irrespective of the considered architecture, we implemented the proposed predictive coding dynamics through a stack of modules called \textit{PCoder}s. The activity of each PCoder $m_{i}$ at time-step $t$ is driven by four terms, as described in the following equation:




\begin{equation}
\label{eq:pc_equation}
    m_{i}(t+1) = \mu m_{i}(t) + \gamma \mathcal{F}_{i}(m_{i-1}(t+1), \theta_{i}^{ff}) + \beta \mathcal{B}_{i+1}(m_{i+1}(t), \theta_{i+1}^{fb}) - \alpha \nabla{\epsilon_{i}(t)},
\end{equation}

\begin{equation}
\label{eq:pc_equation2}
    \epsilon_{i}(t) = \textit{MSE}(\mathcal{B}_{i}(m_{i}(t), \theta_{i}^{fb}), m_{i-1}(t)),
\end{equation}

where $\mathcal{F}_{i}$ computes the feed-forward drive of the $i$th PCoder with parameters $\theta_{i}^{ff}$, and $\mathcal{B}_{i+1}$ computes the feedback drive (prediction) with parameters $\theta_{i+1}^{fb}$ given $m_{i+1}$. The gradient $\nabla{\epsilon_{i}(t)}$ is calculated with respect to the activity of the higher layer ($m_{i}(t)$) as suggested in predictive coding theory.

A specific hyper-parameter modulates each term. First, each PCoder's activity is initialized by a feed-forward pass, i.e., without considering memory or top-down connections, in line with experimental observations in biological visual systems~\cite{vanrullen2001bird, vanrullen2001time}. Then, at successive time-steps, the activity is determined by several terms. First, a memory term, regulated by the $\mu$ hyper-parameter, that retains information from previous time-steps, essentially acting as a time constant. The $\gamma$ and $\beta$ hyper-parameters modulate the feed-forward drive and feedback error terms, which reflect information from the lower and higher layers, respectively. The modulation of the first three terms is normalized, i.e. $\beta + \gamma + \mu = 1$. Lastly, the $\alpha$ hyper-parameter modulates the feed-forward error term, which aims at reducing the prediction-error, i.e. the mean squared error (MSE) between the prediction by a PCoder and the activity of the lower one (or the ``input stimuli'' in case of the first PCoder). As an implementation detail, we multiply $\alpha$ by a scaling factor (see Appendix~\ref{supp:imp_details}) to remove the effect of batch, layer, and (de)convolution kernel size. As postulated by predictive coding formulation, the feedback and feed-forward error terms regulate each PCoder's activity to reduce prediction-errors over time. Importantly, the dynamic described above is equivalent to the one proposed by Rao and Ballard in 1999~\cite{rao1999predictive}, with the only difference being the feed-forward term (for the mathematical proof see~\cite{choksi2021predify}). 

\subsection{Architectures}
\paragraph{Shallow model}
We first implemented a shallow three-layer CNN with two additional dense layers having $120$ and $10$ neurons, respectively. As shown in figure~\ref{fig:architectures}A, the convolutional layers have $12$, $18$ and $24$ channels and a kernel size equal to $5\times5$. Max-pooling operations with stride equal to $2$ were applied from lower to higher layers. In this network, we consider each convolutional layer as a PCoder which predicts the lower one's activity through a bilinear upsampling operation with scale factor equal to $2$, followed by a transposed convolutional layer with window size equal to $3\times3$. The number of channels for the transposed convolution is set in accordance to the prediction target.

\paragraph{Extending to Deep Architectures} Given a very deep architecture, it is not computationally efficient to have every layer predicting the preceding one. Instead, we decided to add PC dynamics to blocks of layers (i.e. each PCoder is a sequence of layers). We took advantage of ``Predify'', a python package introduced in~\cite{choksi2021predify}, that allows to introduce PC dynamics in pre-trained feed-forward networks. In the present paper, we introduce PResNet18 and PEffNetB0 by adding the proposed PC dynamics to feed-forward ResNet18 and EfficientNetB0 architectures, respectively.

To explore more diversity over input images and network depth, we examined PResNet18 and PEffNetB0 on CIFAR100 and ImageNet, respectively. For PEffNetB0 we used the original EfficientNetB0 architecture with pretrained weights on ImageNet as the feed-forward backbone; However, in order to improve ResNet18 performance on small CIFAR100 images, we lowered the kernel size of the first convolutional layer to $3\times3$ and omitted its following max-pooling layer to prevent information loss in early layers.

We implemented the block-wise PC dynamics into ResNet18 and EfficientNetB0 by splitting their layers into five and eight PCoders, respectively (see supplementary section~\ref{supp:netarch}). Regardless of the feed-forward architecture, we used a general procedure to define the feed-forward ($\mathcal{F}$) and feedback ($\mathcal{B}$) drive modules. Assume that there are $n$ blocks of layers in the feed-forward network. Let $y = f_i(x)$ denote the computation done by block $i$ where $x$ and $y$ have the size $(c_{in}, h_{in}, w_{in})$ and $(c_{out}, h_{out}, w_{out})$, respectively. Then, $\mathcal{F}_i$ is $f_i$ and $\mathcal{B}_i$ is a 2D up-scaling operation by the factor of $(h_{in}/h_{out}, w_{in}/w_{out})$ followed by a transposed convolutional layer with $c_{out}$ channels and $3\times3$ window size.

\subsection{Training Parameters}
\paragraph{Supervised feed-forward}
In both shallow and deep models we trained the feed-forward ($\theta_i^{ff}$) and feedback ($\theta_i^{fb}$) parameters separately with different loss functions. First, we trained $\theta_i^{ff}$ to optimize the cross-entropy loss (classification) without using the iterative PC dynamics (i.e., in one forward pass). Accordingly, we used a cross-entropy loss with Stochastic Gradient Descent (SGD) optimizer for the shallow model with  learning rate 0.01 and momentum 0.9. In the case of deep networks, we trained the modified ResNet18 on CIFAR100 training images for 200 epochs using SGD optimizer with initial learning rate 0.1, momentum 0.9, and weight decay 5e-4. We applied learning rate decay factor 0.2 at epochs 60, 120, and 160. For PeffNetB0, we used the pretrained ImageNet model described in~\cite{tan2019efficientnet}.

\paragraph{Unsupervised feedback}
Next, we optimized $\theta_i^{fb}$s with reconstruction objectives, that is the MSE between the activity of PCoders and their top-down reconstruction on the next time-step. This unsupervised approach is akin to a generative process, in which higher layers predict the activity of lower layers, in line with the predictive coding framework. For the shallow network we used an SGD optimizer  with learning rate 0.01 and momentum 0.9. While for both of the deep architectures, we employed Adam~\cite{kingma2014adam} optimizer with learning rate 0.001 and weight decay 5e-4 for 50 epochs.

\paragraph{Supervised feedback}
In the shallow model, we also explored the role of the top-down connections when their parameters are trained for classification rather than reconstruction (as in the previous case). In this case both the $\theta_i^{ff}$ and $\theta_i^{fb}$ are optimized simultaneously for 10 time-steps to minimize the cross-entropy loss. We used an SGD optimizer with learning rate = 0.005 and momentum = 0.9.
Since the learning takes place over time-steps, the network optimizes the weights given the PC dynamics described in equation~\ref{eq:pc_equation}. Importantly, during learning we kept the hyper-parameters values to $\gamma = \beta = \mu = 1/3$ and $\alpha = 0.01$.

\subsection{Training Hyper-Parameters}
After the training of the network's parameters, we froze them (including the statistics of batch normalization layers) and optimized uniquely the hyper-parameters $\gamma$, $\beta$ and $\alpha$ (with $\mu$ constrained to be $1-\beta-\gamma$, see Appendix~\ref{supp:imp_details}). Particularly, we repeated the optimization multiple times with different noise types and levels, to investigate the role of each term given different levels of perturbation. We considered a Cross-Entropy loss function averaged across time-steps. In the shallow model we used an Adam optimizer with learning rate equal to $0.001$, a weight decay equal to 5e-4 and a batch size of 128 images. For each noise type and level, we repeated the experiment with 10 random initializations of each hyper-parameter drawn from the uniform probability distribution in the interval $[0,1]$. We used Adam optimizer with the same weight decay for deep models; however, we employed two separate learning rates equal to $0.01$ for $\gamma$ and $\lambda$, and $0.0001$ for $\alpha$. We set batch-size to $128$ and $16$ for PResNet18 and PEffNetB0, respectively. All the scripts and the trained parameters of the main experiments are available on GitHub\footnote{\url{https://github.com/artipago/Role_of_Feedback_in_Predictive_Coding}}.

\subsection{Stimuli}
The parameters of both the shallow and the deeper networks were trained on clean images, using CIFAR-10, CIFAR-100 and ImageNet. The hyper-parameters were optimized using different levels and types of noise. Regarding the shallow model, we used additive Gaussian and Salt\&Pepper noise, spanning 3 different levels (Gaussian: $\sigma$ = $0.2$, $0.4$ and $0.8$; Salt\&Pepper: pixel percentage = $2\%$, $4\%$ and $8\%$). We used CIFAR100-C, a dataset containing five levels of 19 different corruption types~\cite{hendrycks2019benchmarking} to train PResNet18's hyper-parameters. Finally, in order to train hyper-parameters of the deep PEffNetB0, we used the ImageNet validation set and applied five levels of Gaussian ($\sigma$= $0.5$, $0.75$, $1$, $1.25$, and $1.5$) and Salt\&Pepper (percentage = $5\%$, $10\%$, $15\%$, $20\%$, and $30\%$) noise.

\begin{figure}
    \centering
    \includegraphics[width=0.75\textwidth]{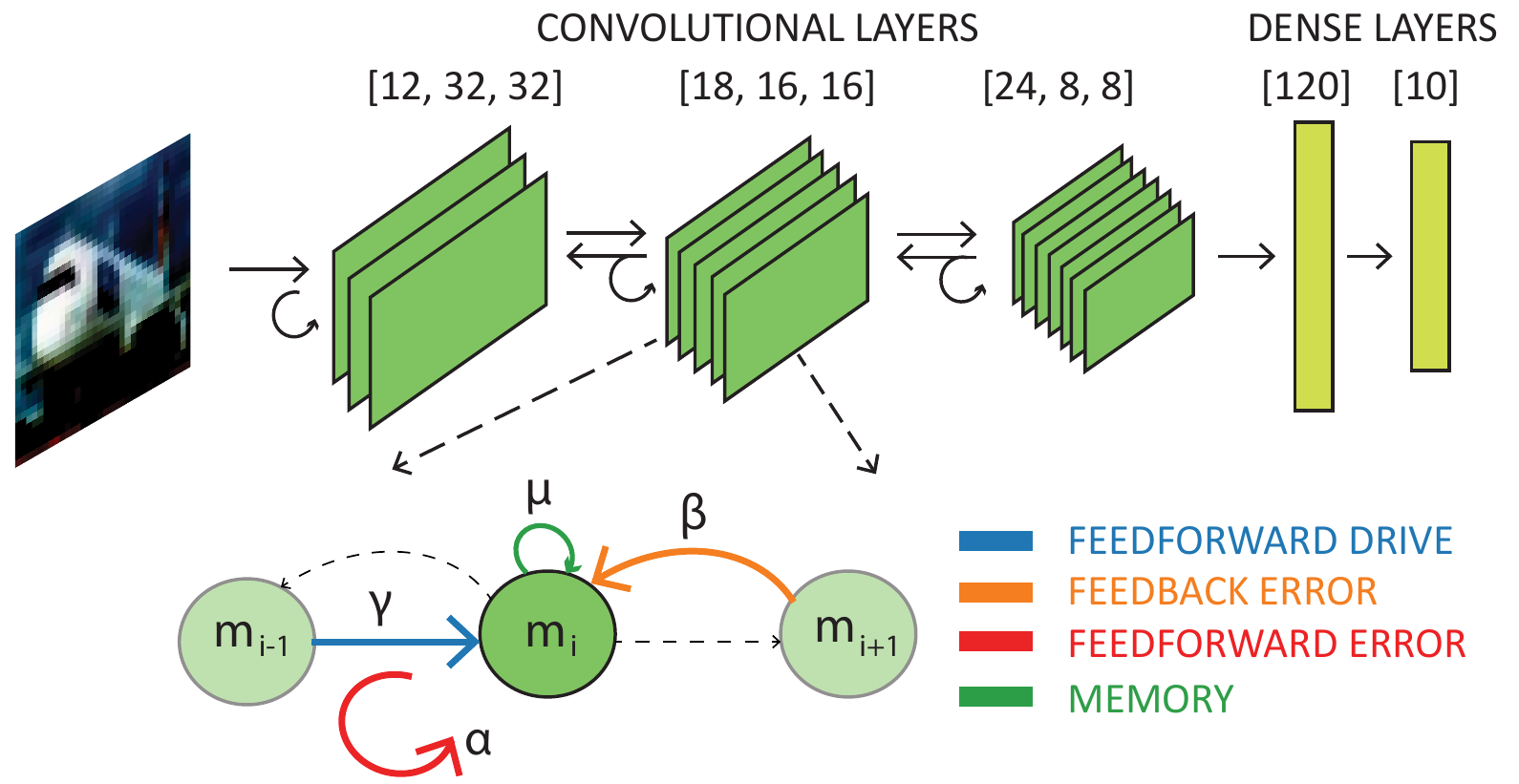}
    \caption{Shallow model architecture and Predictive Coding dynamics. The upper part shows the architecture of the shallow model, composed of three convolutional layers and two fully connected ones. According to Predictive Coding dynamics, the convolutional layers' activity is regulated by four terms, each one modulated by a specific hyper-parameter.}
    \label{fig:architectures}
\end{figure}

\section{Results}
\label{results}

\subsection{Three-Layer Model}
We first tested our hypothesis on a shallow model composed of three convolutional and three dense layers (see panel A of figure~\ref{fig:architectures}). The advantage of choosing a smaller network consists in promptly exploring several approaches before replicating in deeper state-of-the-art networks. Specifically, we investigated the role of each term in equation~(\ref{eq:pc_equation}): 1) when training feedback weights for reconstruction or classification (unsupervised vs supervised), 2) via some ablation simulations, and 3) regarding the robustness to adversarial attacks.  

\subsubsection{Feedback weights: reconstruction vs. classification}
\begin{figure}
    \centering
    \includegraphics[width=0.9\textwidth]{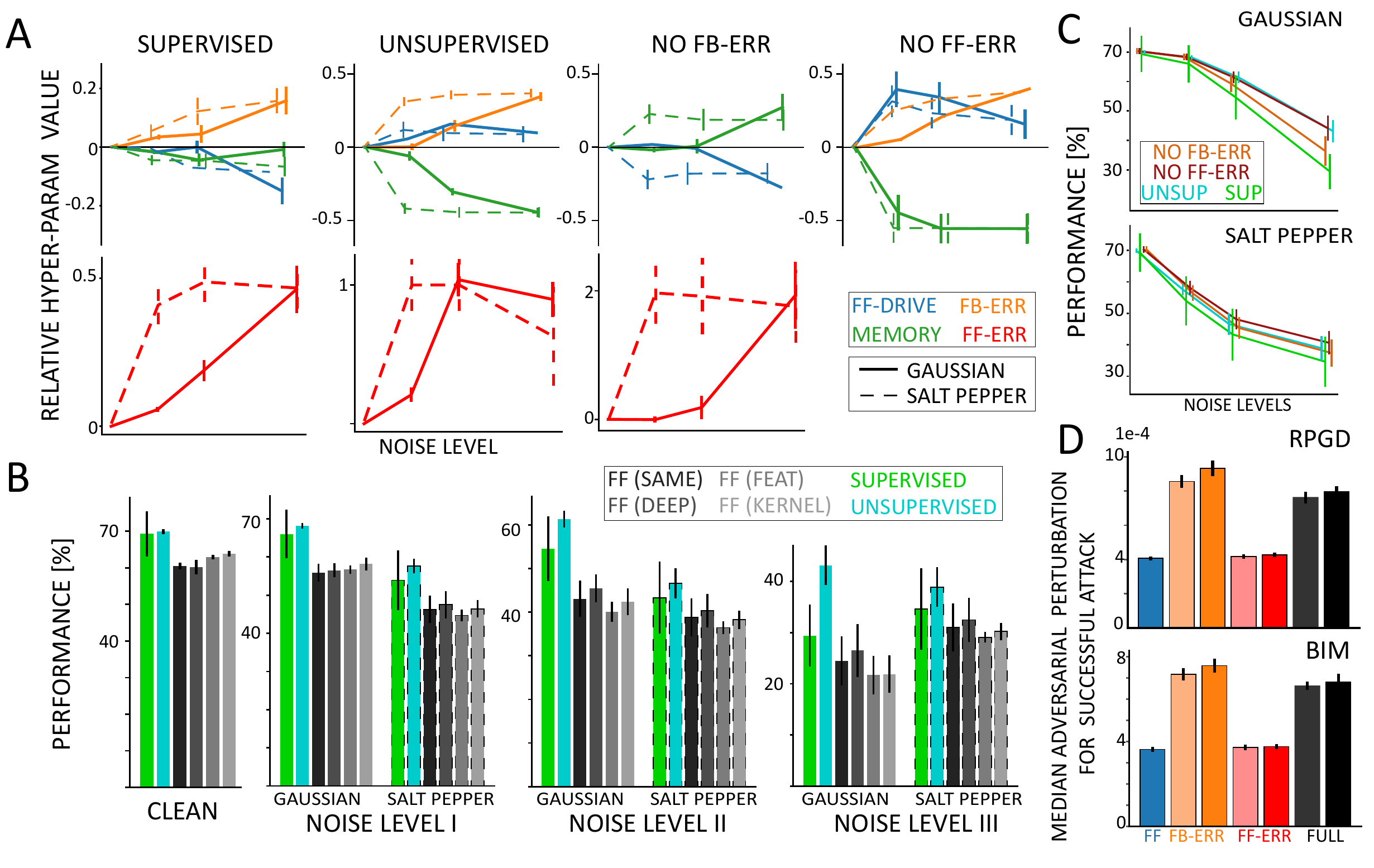}
    \caption{Shallow model results. A) The plots show the hyper-parameters (HPs) value relative to the clean images as a function of the noise levels. Each column shows the relative HPs trained in different conditions: supervised, unsupervised, without feedback error, or forward error. The first row shows the feedback error term, the memory, and the forward drive term, the second row shows the forward error term on a separate scale, for Gaussian (solid line) and Salt\&Pepper (dashed lines) noise. In all conditions, the feedback-error and forward-error terms increase with the noise levels.  B) The models implementing PC dynamics (in green and cyan) perform better than equivalent feed-forward networks, especially when trained in an unsupervised fashion (cyan). C) Performance of the PC models, as a function of the noise levels, measured at the last time-step. Contrasting supervised (SUP) and unsupervised (UNSUP) models reveals the effects of feedback training objective, whereas comparing the ablation models with UNSUP shows the effect of each error term on accuracy. D) The graph shows the median perturbation to obtain a successful attack using different HPs. Orange and red bars have higher feedback and forward error terms (paler colors correspond to smaller error terms), and blue and black bars represent the feed-forward and the full model, respectively. Our simulations reveal that PC models with higher feedback values (orange, black bars) are more robust to adversarial perturbations.}
    \label{fig:shallow_results}
\end{figure}

We first assessed the role of the feedback and each term in equation~(\ref{eq:pc_equation}) when the top-down parameters were optimized for reconstruction. After having trained the forward weights for classification (Supplementary figure~\ref{supfig:shallow_training}A), we trained the feedback weights optimizing the reconstruction loss of each PCoder (figure~\ref{supfig:shallow_training}C). This approach is in line with the PC interpretation, in which top-down connections generate predictions to explain lower layers' activity (i.e., minimize prediction errors, or the reconstruction loss). In this case backward weights are trained in an unsupervised fashion. Once both forward and backward connections were optimized (for classification and reconstruction, respectively), we froze all parameters and trained only the hyper-parameters ($\gamma$, $\beta$ and $\alpha$ in equation~\ref{eq:pc_equation}). As shown in figure~\ref{fig:shallow_results}A, with both Gaussian and Salt\&Pepper noise the hyper-parameter modulating the top-down feedback (i.e., $\beta$ in equation~\ref{eq:pc_equation}) increases as a function of the noise level, supporting the hypothesis that top-down connections are crucial for visual processing in noisy conditions. Remarkably, also $\alpha$, which modulates the amount of bottom-up prediction-error, increases with the noise level for both types of noise. 
Similar results were obtained when training the top-down parameters for classification rather than reconstruction (i.e., supervised approach). As in the unsupervised case, when freezing the parameters and optimizing exclusively the hyper-parameters for different noise levels, we observed an increase of both bottom-up ($\alpha$) and top-down ($\beta$) errors as a function of the noise level. Yet, figure~\ref{fig:shallow_results}C shows that top-down parameters trained for reconstruction proved more robust to noisy images than those trained for classification.
Next, we compared the networks' performance with equivalent forward networks. First, we trained (on clean images) four types of forward networks: either having the same forward architecture as the shallow network (labeled ``same'' in figure~\ref{fig:shallow_results}B, and resulting in a slightly smaller number of parameters), or having a larger number of parameters by increasing either the kernel size, or the number of features, or the layers (labeled ``kernel'', ``feat'' and ``deep'', respectively). As summarized in figure~\ref{fig:shallow_results}B, both networks implementing predictive coding dynamics (in cyan and green in the figure) perform systematically better than all the forward networks, irrespective of the noise type and level. This result demonstrates that feedback connections, and specifically predictive coding dynamics, improve network robustness to noise. 

\subsubsection{Ablation studies}
We then investigated how selectively removing the top-down or the bottom-up error term influences the results. Importantly, we focused specifically on the unsupervised network, whose top-down parameters are trained for reconstruction, and that better represents the PC dynamics. As shown in figure~\ref{fig:shallow_results}A, when removing the top-down error term, the forward error hyper-parameter increases with the noise levels and doubles its value as compared to the full model (labeled ``unsupervised" in the figure). On the other hand, when removing the forward error term, we observed an increase of the feedback term with the noise levels, as in the full model. Concerning the networks performance, figure~\ref{fig:shallow_results}C reveals that removing the top-down feedback degrades the accuracy with higher noise levels (especially with Gaussian noise), confirming the conclusion that top-down feedback plays a crucial role in the processing of degraded images.

\subsubsection{Adversarial attacks}
To further confirm the hypothesis that top-down feedback is important for robustness, we froze the networks (feedforward, full predictive coding, or ablated networks) with manual configurations of the hyper-parameters and then tested their robustness against targeted $L_{\infty}$ Random Projected Gradient Descent (RPGD)~\cite{madry2017towards} and Basic Iterative Method (BIM)~\cite{goodfellow2014explaining} attacks, after unrolling them for 10 time-steps to keep their depths constant. We use Foolbox API 2.4.0~\cite{rauber2017foolbox} and measure the median perturbation required to successfully fool the networks. As shown in figure~\ref{fig:shallow_results}D, we observe that networks with higher top-down feedback (two orange bars in the figure have $\alpha = 0$ and $\gamma = \beta = \mu = 0.33$; and $\beta = 0.5, \gamma = 0.3, \mu = 0.2$, respectively) reveal better robustness to the attacks as compared to the equivalent forward network (in blue in the figure, with $\gamma=1$ and all other hyper-parameters set to zero). Interestingly, a forward network leveraging only the feed-forward error shows a similar (lack of) robustness to the attack as the forward network (in red in the figure, both networks having $\gamma=1$, and $\alpha = 1$ and $\alpha = 2$, respectively; all other hyper-parameters set to zero). Additionally, adding the feed-forward error to the model with top-down connections, slightly reduces its robustness (black bars in the picture, both networks with $\alpha = 1$ and $\gamma = 0.3$, while $\beta = \mu = 0.33$ and  $\beta =0.5$ $\mu = 0.2$, respectively). These results confirm that top-down connections are useful for adversarial robustness (as shown on a different dataset with a different PC implementation by Huang and colleagues~\cite{huang2020neural}), but also suggest that feedforward error correction does not help adversarial robustness. This is likely because the feedforward prediction errors emphasize the input perturbation, which the generative feedback was not trained to account for.

\subsection{Deep Models}

\subsubsection{Shared hyper-parameters}
Similar to the three-layer network, we examined PResNet18 with a single set of $\alpha, \beta$, and $\gamma$, that is shared between all the PCoders. In this experiment, we followed the unsupervised training approach explained for the three-layer network using the CIFAR100 dataset. After having optimized the top-down connections for reconstruction, we froze the weights and trained the hyper-parameters to minimize the average cross-entropy loss over five time-steps. We performed this optimization independently on each noise type and noise level of the CIFAR100-C dataset. 

\begin{figure}
    \centering
    \includegraphics[width=0.85\textwidth]{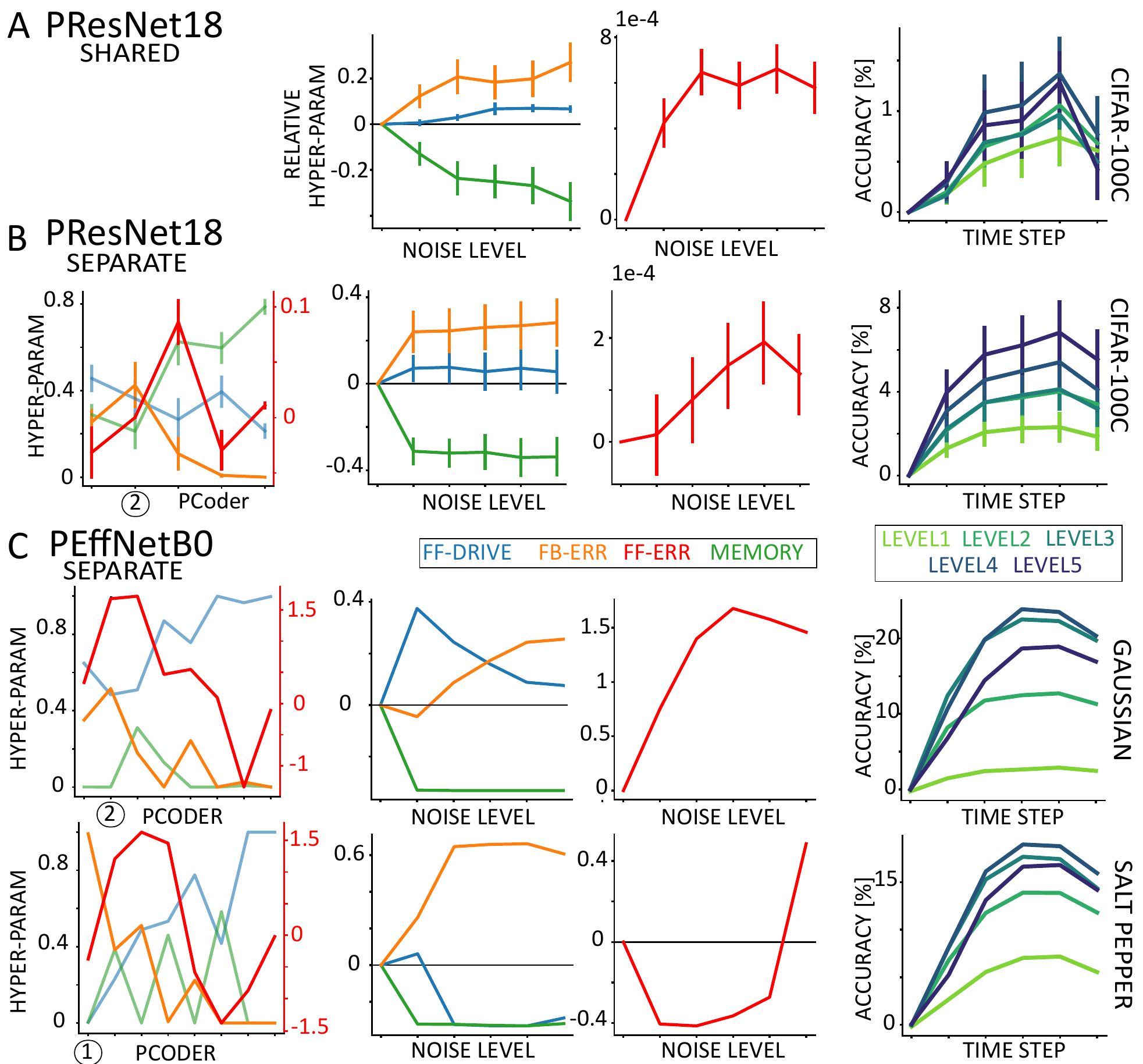}
    \caption{Values of hyper-parameters and accuracy of the deep predictive coding networks. (A) PResNet18 with shared hyper-parameters that are trained on CIFAR100-C images. (B) PResNet18 and (C) PEffNetB0 with separate hyper-parameters that are trained respectively on CIFAR100-C and ImageNet under Gaussian and Salt\&Pepper noise. Plots in the first column show the hyper-parameters as a function of PCoders under medium noise level. The circles indicate PCoders with maximum feedback error. In middle columns, relative values of hyper-parameters are plotted across noise levels. In case of separate hyper-parameters, the PCoder with maximum feedback error is shown. Accuracy change for each noise level is depicted in the last column. Error bars show  standard error of the mean (SEM) over 19 CIFAR100-C noise types. In all cases, the networks achieve accuracy gain by utilizing more feedback and forward error as the noise severity increases. See supplementary Figures~\ref{supfig:presnet_shared_accuracy}-\ref{supfig:peffnetb0_separate_hyperparameters_saltpepper} for the absolute values of hyper-parameters and changes in recognition accuracy per noise type and level.}
    \label{fig:deep_results}
\end{figure}

Figure~\ref{fig:deep_results}A shows the average hyper-parameter values across all 19 noise types relative to those learned using ``clean'' images. Confirming the results of the shallow model, we observed that the roles of feedback and feed-forward error become more crucial as the noise level increases. Importantly, the average accuracy change across time-steps reveals a very robust (but marginal) improvement with respect to the feed-forward ResNet18 for all levels of noise. Remarkably, the importance of feedback connections shines more as the noise severity increases. 

\subsubsection{Separate hyper-parameters}
Encouraged by the results in the ``shared'' approach described above, we decided to provide each PCoder with a separate set of hyper-parameters. Our reasoning was that different stages of the hierarchical visual processing would benefit differently from the combination of top-down and bottom-up information, thus granting to the network more flexibility in accounting for different representations across different layers.

As in the previous experiment, we trained PResNet18's hyper-parameters on CIFAR100-C images. Moreover, in order to validate our previous results on a more complex dataset, we trained PEffNetB0's hyper-parameters on the ImageNet2012 validation set for five levels of Gaussian and Salt\&Pepper noises.

Introducing a separate set of hyper-parameters in each PCoder resulted in a very significant boost in recognition accuracy of both networks, under all conditions. As illustrated in the last column of Figure~\ref{fig:deep_results}B, PResNet18 consistently improved the recognition accuracy across time-steps on all noise types and levels, revealing an average improvement around $6\%$ in the most noisy condition.  Remarkably, we could replicate these results using the deeper network PEffNetB0 with eight PCoders. As shown in Figure~\ref{fig:deep_results}C, PC dynamics with different hyper-parameters per PCoder yielded an impressive increase in accuracy above $20\%$ and above $15\%$ in the worst condition of Gaussian and Salt\&Pepper noise, respectively. 

We then investigated the trend of hyper-parameters across PCoders. This analysis shed some light on the role of the hyper-parameters as a function of their hierarchical stage in the network. Remarkably, we obtained very consistent results on both networks, and across different noise types. The first column in panels B and C of Figure~\ref{fig:deep_results} shows the values of hyper-parameters as a function of PCoders for the medium noise level (level 3, results don't change across noise levels, see supplementary figures~\ref{supfig:presnet_separate_hyperparameters}-\ref{supfig:peffnetb0_separate_hyperparameters_saltpepper}). Regardless of the considered model, we found that the PCoder with the largest amount of feedback error hyper-parameter (indicated by a circle in the figure), is consistently situated at the lower layers of the network, whereas the feedback tends to zero at higher layers. This suggests that the beneficial effects of top-down connections are best achieved at lower layers of the visual hierarchy, where high-level expectations shape low level features to maximize the final classification. 

In addition, the second and third columns of Figure~\ref{fig:deep_results}B, C confirmed our previous results, revealing how the feedback-error term increases as a function of the noise levels in the PCoder with its highest values (i.e., the second for PResNet18, and either the first or the second in PEffNetB0, depending on the noise type). This result confirms once again the hypothesis that robust object recognition requires more top-down influence (i.e., feedback and feed-forward error terms) as the level of noise increases.

\section{Discussion}
\label{discussion}
\subsection{Summary of the Results}
Starting from an established framework in Neuroscience, namely Predictive Coding (PC), we investigated the role of top-down feedback connections in models of vision. The significance of our work spans across Neuroscience and machine learning, contributing substantially to both fields. First, our results demonstrated how predictive coding dynamics increase the network's robustness to various types of noisy stimuli compared to equivalent feed-forward networks. Additionally, systematic optimization of hyper-parameters revealed how the feedback contribution increases with the noise severity, especially in the early stages of the network, providing important information about the role of top-down processes in visual processing. Compared with prior studies, one original aspect of our approach is our empirical procedure, in which we let the optimization process converge to the optimal solution in each noise level.

\subsection{Previous Work}
Previous studies explored the supervised approach to train feedback connections for classification rather than reconstruction objectives. Feedback Networks~\cite{zamir2017feedback} introduced top-down and temporal skip connections in a recurrent convolutional module, demonstrating an increase in performance followed by improvements in early features representation, taxonomic predictions, and curriculum learning. Similarly, Nayebi and colleagues \cite{nayebi2018task} proposed a ConvRNN architecture, incorporating gating and skip connections, which significantly improved object recognition performance. Considering models advocating more explicitly for biological plausibility, Linsley and colleagues~\cite{linsley2018learning} suggested another recurrent vision model, equipped with horizontal and gated recurrent units (hGRU). Its performance improves specifically in recognition tasks involving long-range spatial dependencies. Supported by experimental studies~\cite{kar2019evidence}, Kubilius and colleagues also proposed a brain-inspired architecture named CorNet, which includes feedback and skip connections. Interestingly, it reveals high neural similarity to cortical visual areas such as V4 and IT~\cite{kubilius2018cornet}.

In the PC domain, Chalasani and Principe~\cite{chalasani2013deep} proposed a hierarchical, generative model based on PC dynamics, including context-sensitive priors on the latent representations. Their architecture demonstrated how top-down connections from higher layers are instrumental in solving lower layers ambiguities, providing some noise robustness. The model proposed in~\cite{wen2018deep} is the closest one to ours. Despite following PC dynamics and the principal similarities, their model presents some critical limitations. More specifically, all weights are trained for object recognition only at the last time step, resulting in a biologically implausible behavior, in which near-chance performance is observed until the final iteration. A more in-depth comparison between this work and our proposed method is presented in~\cite{choksi2021predify}. Finally, Huang and colleagues~\cite{huang2020neural} implemented unsupervised feedback connections by optimizing for ``self consistency'' between the input image features, latent variables and label distribution. Despite a different dynamics, PC principles inspired their implementation, which also provided some robustness against gradient-based adversarial attacks on Fashion-MNIST and CIFAR10. 

\subsection{Insights for and from Neuroscience}
It is possible to characterize the role of top-down feedback either as an unsupervised, generative process which predicts lower layers' activities, or as a supervised, discriminative process to optimize classification. Besides being more biologically plausible, our simulations with the shallow model revealed that the unsupervised approach is more robust to noise than the supervised one, as shown in figure~\ref{fig:shallow_results}B. However, when trained with supervision, feedback connections do not converge to the unsupervised solution, as shown in figure~\ref{supfig:shallow_training}C which compares the reconstruction errors in shallow models optimized for classification (supervised) or reconstruction (unsupervised). 


Interestingly, when we independently optimized each PCoder in deeper networks (roughly equivalent to different brain regions across the hierarchy of visual processes), we observed consistently higher modulation of top-down activity in lower regions, and relatively less top-down feedback in higher areas. Choksi and Mozafari et al.~\cite{choksi2021predify} further demonstrates that the proposed biologically-inspired feedback dynamics iteratively project the noisy inputs towards the learned data manifold, similar to previous studies using different approaches~\cite{meng2017magnet,shen2017ape,samangouei2018defense,jalal2017robust}. Future research may test this prediction directly in biological brains by recording the top-down cortical activity at different stages of the visual hierarchy, and validate the hypothesis that early brain regions benefit the most from top-down feedback during visual perception in noisy conditions.

Our results demonstrated how top-down and bottom-up processes influence perception in different challenging conditions. However, how does the brain modulate each term’s contribution (i.e., each hyper-parameter) during natural vision? Attention mechanisms may be responsible for the regulation of top-down processes by increasing feedback response during noisy conditions~\cite{baluch2011mechanisms, feldman2010attention}. Accordingly, it could be possible to envision a model inspired by current transformer architectures where a biologically plausible attention system modulates hyper-parameters based on input features or top-down expectations~\cite{vanrullen2021gattanet}. Expectation is another important process that modulates top-down feedback in the human brain~\cite{de2018expectations, summerfield2014expectation}. In our model, the forward pass initializes the activity in each layer based on the first processing of the input (i.e., without the recurrent PC dynamic). However, it is possible to initialize the network's activity based on top-down beliefs, according to PC dynamics: the last layer of the hierarchy encodes the predictions of the expected input (i.e., a given class in a classification dataset), and propagates such predictions to initialize the activity of lower layers, similarly to the brain processes involved in sensory expectations~\cite{summerfield2009expectation, kok2015predictive}. Future work could explore how such expectations may influence the network behavior and accuracy.

\section*{Broader Impact}

Despite its outstanding achievements, artificial intelligence (AI) revealed significant reliability limitations when tested in challenging conditions. Addressing these concerns is becoming a crucial goal for the scientific community, as AI is gaining an important place in our daily lives. In this work, we leverage an established framework in Neuroscience –namely Predictive Coding- to address this problem and investigate the role of feedback in robust visual processing. Our results suggested that inspiration from the human brain can be beneficial for artificial vision, and our model provides a remarkable tool to study the visual system in biological brains~\cite{pang2021predictive}. On the one hand brain-inspired approaches can boost artificial sensory processes in ecological (noisy) situations. On the other hand, we are aware of the possible nefarious use of human-like artificial systems, and we encourage researchers and policymakers to consider these issues and their societal implications.

\begin{ack}
RV is supported by an ANITI (Artificial and Natural Intelligence Toulouse Institute) Research Chair (grant ANR-19-PI3A-0004), and two ANR grants AI-REPS (ANR-18-CE37-0007-01) and OSCI-DEEP (ANR-19-NEUC-0004). 
\end{ack}

\bibliographystyle{unsrt}
\bibliography{ref}

\clearpage

\appendix

\section{Appendix}
\label{appendix}

\subsection{Deep Network Architectures}\label{supp:netarch}
In this part, we explain how we split each of the ResNet18 and EfficientNetB0 into blocks of layers and converted them into PCoders.

We used a modified ResNet18 architecture that works better with 32x32 images from the CIFAR100 dataset. ResNet18 is a sequence of residual blocks, each of which consists of a sequence of convolution, batch normalization, and ReLU layers. Due to the residual connections around each block, we chose to never split them into multiple PCoders. However, a single PCoder may contain more than one residual blocks in its feedforward module ($\mathcal{F}$). More precisely, we split the modified ResNet18 into 5 PCoders. PCoder1 contains the first Convolution and Batch Normalization layers. PCoder2 to PCoder5 contain two consecutive residual blocks each. 

In the case of PEffNetB0, we used the PyTorch implementation of EfficientNetB0 provided in \url{https://github.com/rwightman/pytorch-image-models}. In this implementation, EfficientNetB0 is split into eight blocks of layers (considering the first convolution and batch normalization layers as a separate block). Except for the classification block, we converted each block into a PCoder.

Please see Table~\ref{suptab:netarc} for more details on deep predictive coding architectures and their PCoders.

\begin{table}[H]

    \scriptsize
    \centering
    \caption{Architectures of PResNet18 and PEffNetB0. Conv (channel, size, stride), Deconv (channel, size, stride), Upsample (scale\_factor), BN is BatchNorm, $[\ ]_+$ is ReLU, and $[\ ]_*$ is SiLU. EfficientBlock corresponds to each block in the PyTorch implementation of EfficientNetB0. See Table~\ref{suptab:basicblock} for the structure of ResNet BasicBlocks}
    \label{suptab:netarc}
    \begin{tabular}{|c|c|c||c|c|}
        \hline
        & \multicolumn{2}{|c||}{PResNet18} & \multicolumn{2}{|c|}{PEffNetB0}\\
        & \multicolumn{2}{|c||}{Input Size: 3x32x32} & \multicolumn{2}{|c|}{Input Size: 3x224x224}\\
        \cline{2-5}
        & $\mathcal{F}_i$ & $\mathcal{B}_i$ & $\mathcal{F}_i$ & $\mathcal{B}_i$\\
        \hline
        PCoder1 &
        \begin{tabular}{c}
             $[$BN (Conv (64, 3, 1))$]_+$
        \end{tabular}
        &
        \begin{tabular}{c}
             Deconv (3, 3, 1)
        \end{tabular}
        &
        \begin{tabular}{c}
             $[$BN (Conv (32, 3, 2))$]_*$
        \end{tabular}
        &
        \begin{tabular}{c}
             Upsample (2)\\
             Deconv (3, 3, 1)
        \end{tabular}
        \\
        \hline
        PCoder2 &
        \begin{tabular}{c}
             $[$BasicBlock (64, 3, 1)$]_+$ \\
             $[$BasicBlock (64, 3, 1)$]_+$
        \end{tabular}
        &
        \begin{tabular}{c}
             Deconv (64, 3, 1)
        \end{tabular}
        &
        \begin{tabular}{c}
             EfficientBlock0
        \end{tabular}
        &
        \begin{tabular}{c}
             Deconv (32, 3, 1)
        \end{tabular}
        \\
        \hline
        PCoder3 &
        \begin{tabular}{c}
             $[$BasicBlock (128, 3, 2)$]_+$ \\
             $[$BasicBlock (128, 3, 1)$]_+$
        \end{tabular}
        &
        \begin{tabular}{c}
             Upsample (2)\\
             Deconv (64, 3, 1)
        \end{tabular}
        &
        \begin{tabular}{c}
             EfficientBlock1
        \end{tabular}
        &
        \begin{tabular}{c}
             Upsample (2)\\
             Deconv (16, 3, 1)
        \end{tabular}
        \\
        \hline
        PCoder4 &
        \begin{tabular}{c}
             $[$BasicBlock (256, 3, 2)$]_+$ \\
             $[$BasicBlock (256, 3, 1)$]_+$
        \end{tabular}
        &
        \begin{tabular}{c}
             Upsample (2)\\
             Deconv (128, 3, 1)
        \end{tabular}
        &
        \begin{tabular}{c}
             EfficientBlock2
        \end{tabular}
        &
        \begin{tabular}{c}
             Upsample (2)\\
             Deconv (24, 3, 1)
        \end{tabular}
        \\
        \hline
        PCoder5 &
        \begin{tabular}{c}
             $[$BasicBlock (512, 3, 2)$]_+$ \\
             $[$BasicBlock (512, 3, 1)$]_+$
        \end{tabular}
        &
        \begin{tabular}{c}
             Upsample (2)\\
             Deconv (256, 3, 1)
        \end{tabular}
        &
        \begin{tabular}{c}
             EfficientBlock3
        \end{tabular}
        &
        \begin{tabular}{c}
             Upsample (2)\\
             Deconv (40, 3, 1)
        \end{tabular}
        \\
        \hline
        PCoder6 &
        -
        &
        -
        &
        \begin{tabular}{c}
             EfficientBlock4
        \end{tabular}
        &
        \begin{tabular}{c}
             Deconv (80, 3, 1)
        \end{tabular}
        \\
        \hline
        PCoder7 &
        -
        &
        -
        &
        \begin{tabular}{c}
             EfficientBlock5
        \end{tabular}
        &
        \begin{tabular}{c}
             Upsample (2)\\
             Deconv (112, 3, 1)
        \end{tabular}
        \\
        \hline
        PCoder8 &
        -
        &
        -
        &
        \begin{tabular}{c}
             EfficientBlock6
        \end{tabular}
        &
        \begin{tabular}{c}
             Deconv (192, 3, 1)
        \end{tabular}
        \\
        \hline
    \end{tabular}
\end{table}

\begin{table}[H]
    \scriptsize
    \centering
    \caption{Architecture of BasicBlock($i$, $j$, $k$). Each BasicBlock is a residual block where the input is added to the output of the block. When $k\neq1$, the input passes through a Conv($i$,1,2) and a BatchNorm before being added to the output.}
    \label{suptab:basicblock}
    \begin{tabular}{|c|}
        \hline
        BasicBlock ($i$, $j$, $k$)\\
        \hline
        $[$BN (Conv ($i$, $j$, $k$))$]_+$\\
        BN (Conv ($i$, $j$, 1))\\
        \hline
    \end{tabular}
\end{table}

\subsection{Implementation Details}\label{supp:imp_details}
\paragraph{Training Hyper-Parameters}
Since $\mu$, $\gamma$, and $\lambda$ should satisfy the constraint $\mu+\gamma+\lambda=1$, independently optimizing them with backpropagation leads to invalid values. To solve this issue, we made use of auxiliary parameters. Precisely, let $\mu_{aux}$, $\gamma_{aux}$, and $\lambda_{aux}$ denote three auxiliary parameters. Then, we compute $\mu$, $\gamma$, and $\lambda$ as follows:

\begin{equation}
    \mu = \frac{\sigma(\mu_{aux})}{\sigma(\mu_{aux}) + \sigma(\gamma_{aux})+\sigma(\beta_{aux})},
\end{equation}
\begin{equation}
    \gamma = \frac{\sigma(\gamma_{aux})}{\sigma(\mu_{aux}) + \sigma(\gamma_{aux})+\sigma(\beta_{aux})},
\end{equation}
\begin{equation}
    \beta = \frac{\sigma(\beta_{aux})}{\sigma(\mu_{aux}) + \sigma(\gamma_{aux})+\sigma(\beta_{aux})},
\end{equation}
where
\begin{equation}
    \sigma(x) = \frac{1}{1 + \exp(-x)}.
\end{equation}
While the auxiliary parameters can take on any real value, the corresponding hyper-parameters are thus constrained between 0 and 1, summing to 1.

\paragraph{Gradient Scaling}

In our dynamics, the error ($\epsilon_i$) is defined as a scalar quantity whose gradient is taken with respect to the activation of the higher layer ($m_i$). That is,

\begin{equation}
    \epsilon_i = \frac{1}{K}\sum_k^K(m_{i-1}^k - p_{i-1}^k)^2
\end{equation}

where $p_{i-1}$ ($= \mathcal{B}(m_{i},\theta^{fb}_{i+1}) $) represents the prediction made for $m_{i-1}$  and K represents the number of elements in $m_{i-1}$ ( = channels*width*height). Thus, the error-correction term at position $j$ itself becomes,

\begin{align}
\frac{\partial\epsilon_{i}}{\partial m_i^j} & = \label{kceq6} \frac{1}{K}\sum_k^K\frac{\partial(m^k_{i-1} - p^k_{i-1})^2}{\partial m^j_i}
\end{align}

Equation~\ref{kceq6} highlights how the dimensionality of the prediction (equivalently the error term) affects the gradients, scaling them with a factor K that can differ across layers by orders of magnitude. This effect is worsened for CNNs where the gradients outside of the receptive field (of size $C$) of element $m^j_i$ will be zero,

\begin{align}
    \sum_k^K\frac{\partial(m^k_{i-1} - p^k_{i-1})^2}{\partial m^j_i} = \sum_k^C\frac{\partial(m^k_{i-1} - p^k_{i-1})^2}{\partial m^j_i}
\end{align}

To counteract this, we apply a layer-specific scaling factor to the error gradients. Assuming that the partial derivative of the error for each pair of connected neurons $i,j$ is i.i.d normally distributed around 0 :
\begin{align}
    \frac{\partial(m^k_{i-1} - p^k_{i-1})^2}{\partial m^j_i} \sim \mathcal{N}(0,\sigma^2)
\end{align}

It can be shown that, 

\begin{align}
\label{eq9}
    \frac{\partial \epsilon_{i}}{\partial m^j_i} = \frac{1}{K}\sum_k^C\frac{\partial(m^k_{i-1} - p^k_{i-1})^2}{\partial m^j_i} \sim \mathcal{N}(0,\frac{C\sigma^2}{K^2})
\end{align}

Equation~\ref{eq9} provides a way to, at least partly, counteract the effect of the dimensionality for our gradient. We multiply the gradient by a factor of $\sqrt{K^2/C}$ to scale them and apply a more meaningful step size for correcting the errors.

\paragraph{Execution Time}
We tested the shallow models and the deeper networks on different machines. All the simulations of the shallow models, including the 10 different initializations, took approximately 4 days using 1 GPU Nvidia GTX 1080Ti with 11Gb. The training of PResNet18 on CIFAR-C took approximatively 2 weeks, whereas PEffNetB0 was trained in ~6 days, using a machine equipped with 1 GPU Nvidia TitanV with 12Gb.

\subsection{Supplementary Figures}

In this section, we present the supplementary figures which complement the main text. The first three figures (fig. 4-6) show the full results of the shallow model: the value of its hyper-parameters (figure 4), the results concerning the training of its parameters (figure 5), and the accuracy over time steps (figure 6). Figures 7 and 8 show the relative accuracy for PResNet18 on all noise levels in CFIAR100-C, using shared or separate hyperparameters, respectively. The last four figures (fig 9-12) report all the hyper-parameters values for PResNet18 and PEffNetB0, considering different noise levels and -in the case of separate hyper-parameters- each PCoder.

\begin{figure}[H]
    \centering
    \includegraphics[width=0.75\textwidth]{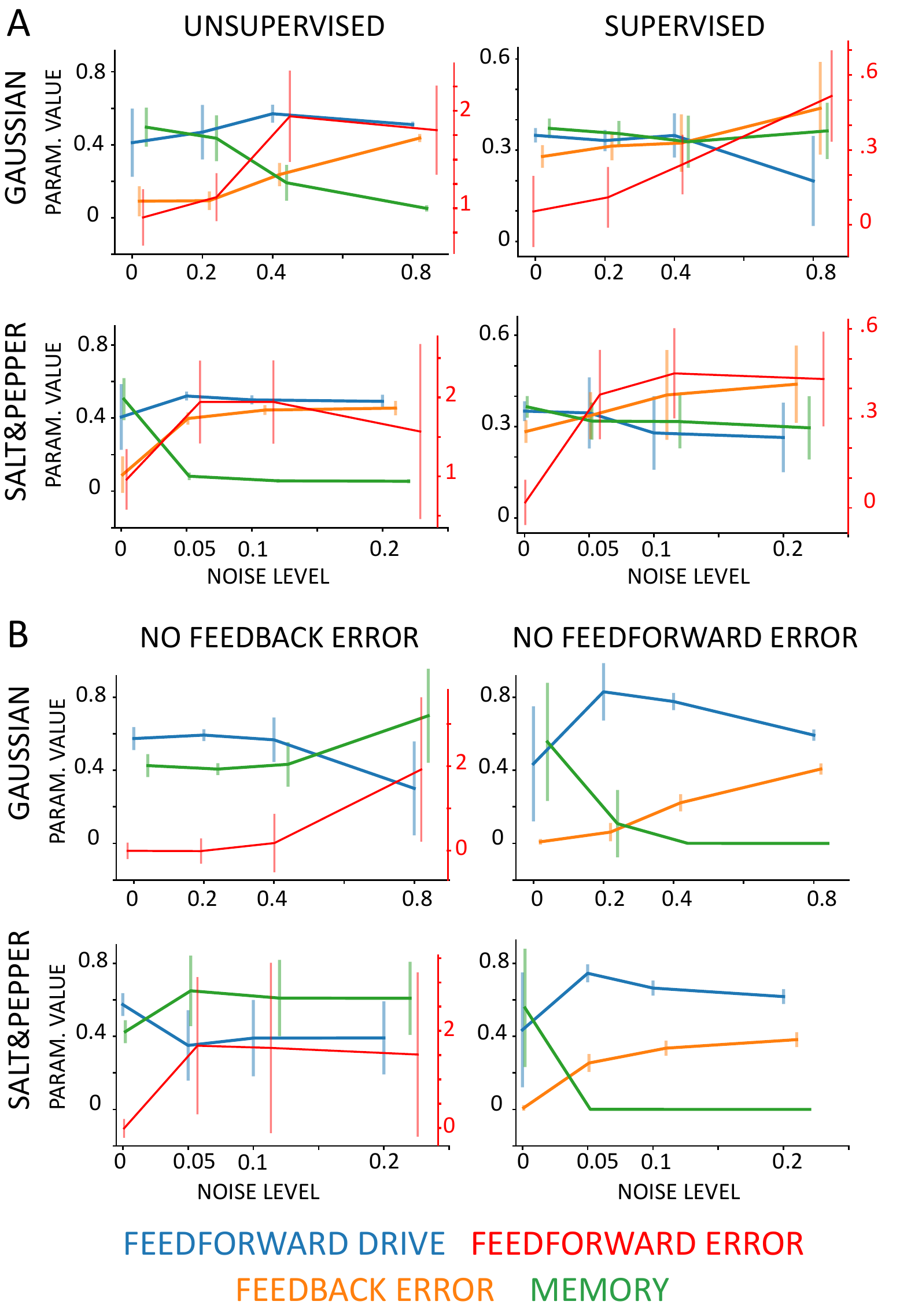}
    \caption{Hyper-parameters for the shallow model. A) Each subplot shows the hyper-parameter values for the unsupervised and supervised networks (left and right column, respectively) for gaussian and salt\&pepper noise (first and second row, respectively). The color code is consistent with the main figure and indicated at the bottom of the figure. The feedforward-error refers to its own y-axis shown to the right of each panel. B) Same as in A but for the ablation models, in which either the feedback-error (left column) or the feedforward-error (right column) were removed.}
    \label{supfig:hyperParameters}
\end{figure}

\begin{figure}[H]
    \centering
    \includegraphics[width=0.75\textwidth]{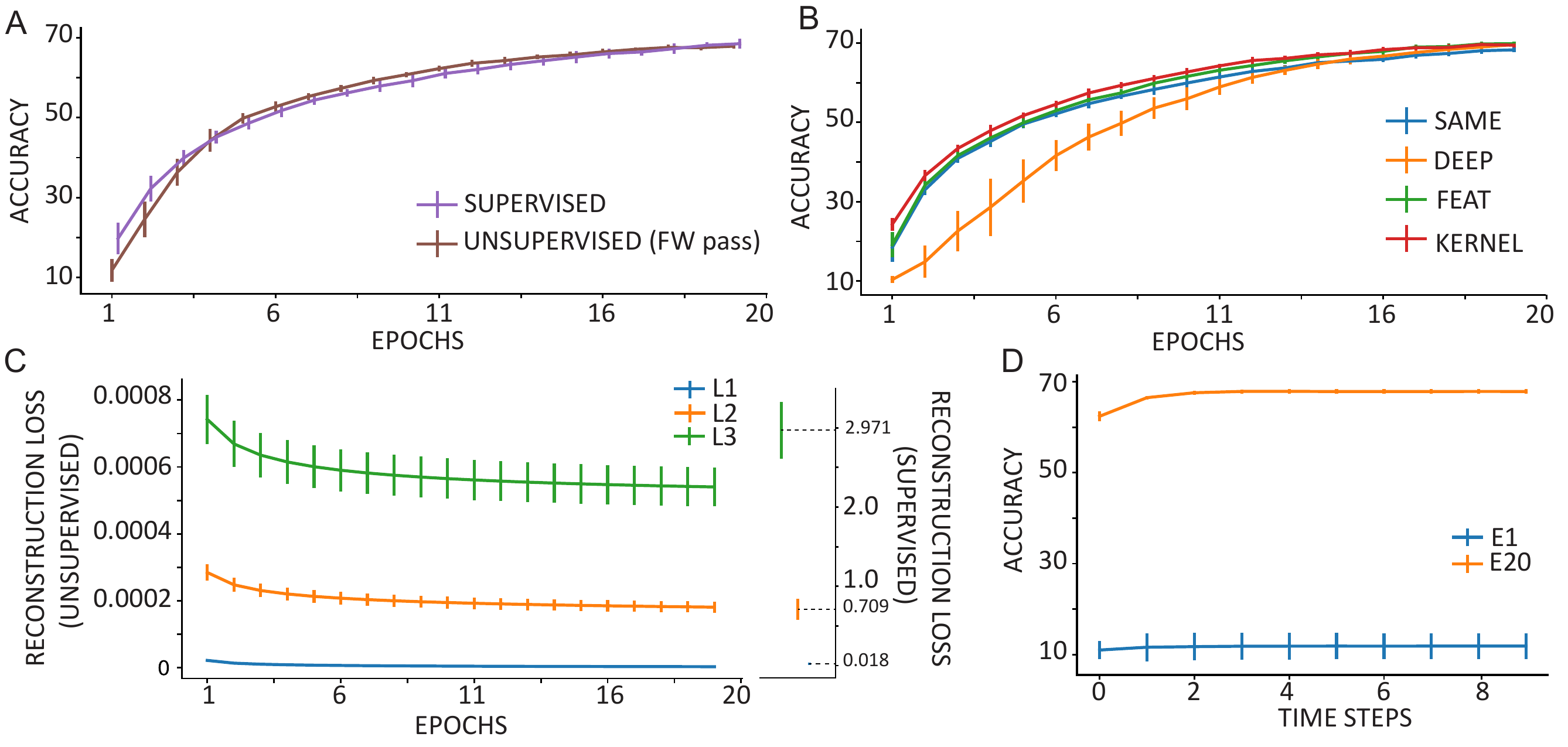}
    \caption{Shallow models training. A) Networks' accuracy during the training. In the supervised case the hyper-parameters were kept fixed, whereas in the unsupervised case we report the accuracy of the forward pass only. B) Training accuracy of the forward networks. C) Reconstruction loss for each Pcoder in the shallow model. For comparison, we report to the right the reconstruction loss in the supervised network (note that in this case the network was not trained for reconstruction). D) Accuracy of the supervised network as a function of the time-steps in the first and last block of training.}
    \label{supfig:shallow_training}
\end{figure}

\begin{figure}[H]
    \centering
    \includegraphics[width=0.75\textwidth]{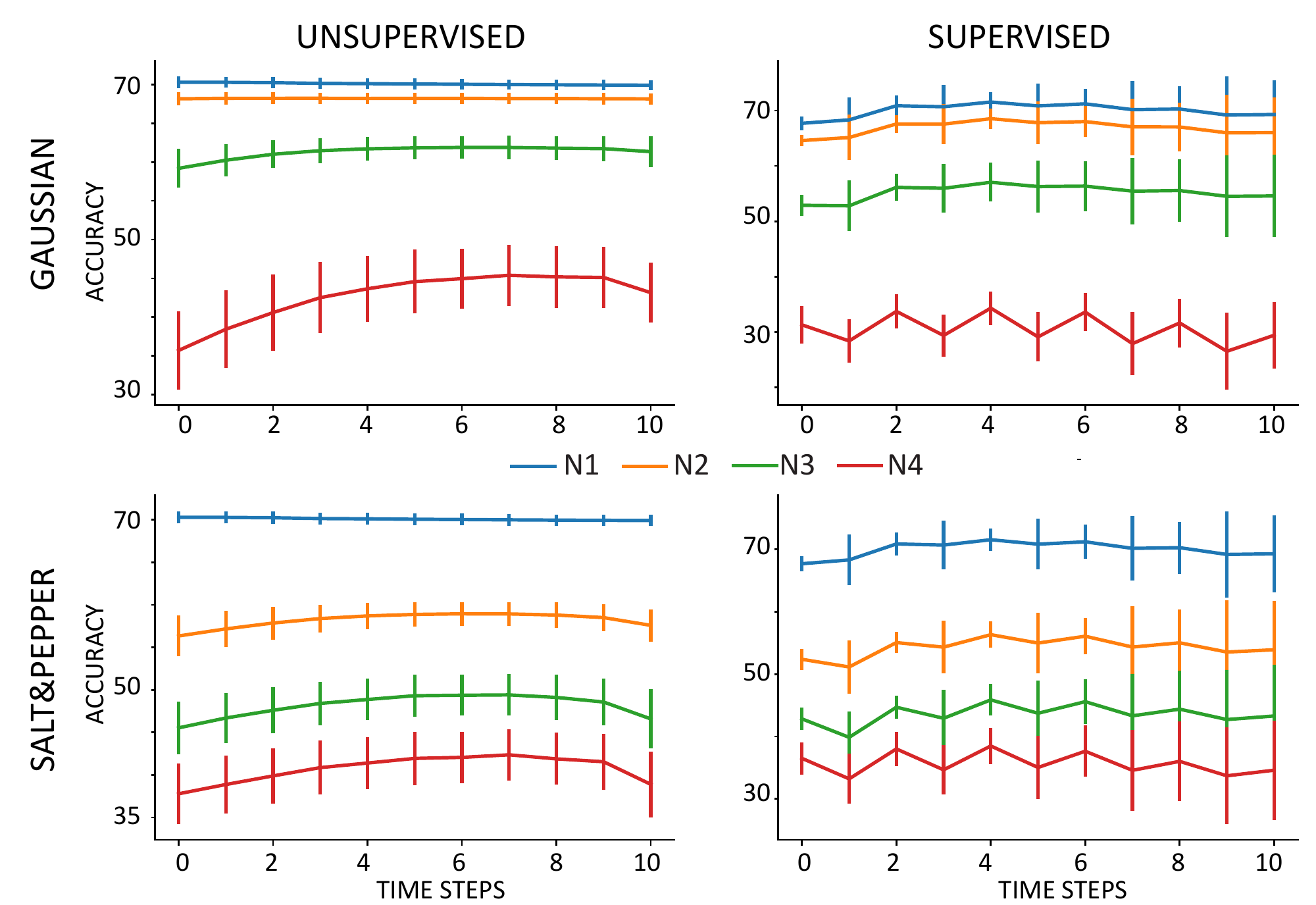}
    \caption{Accuracy over time steps of both shallow models trained for reconstruction (left column) or classification (right column). The colors represent different noise levels for Gaussian (first row) and Salt\&pepper noise (second row).}
    \label{supfig:shallow_accuracy}
\end{figure}

\begin{figure}[H]
    \centering
    \includegraphics[width=\textwidth]{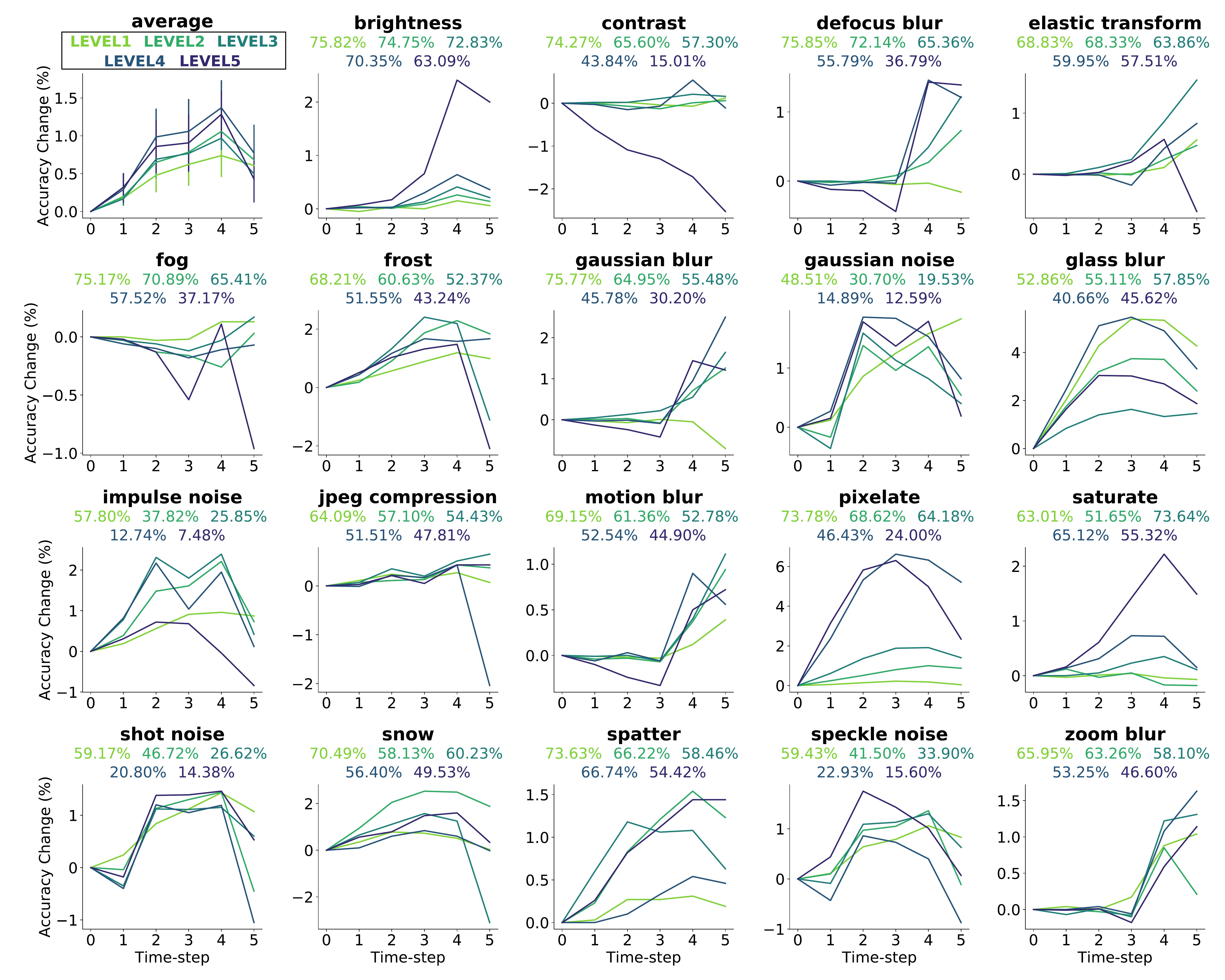}
    \caption{Recognition accuracy of PResNet18 with shared hyper-parameters on each of the CIFAR100-C noise types and levels. Each plot shows the change in accuracy with respect to the feedforward baseline (i.e. ResNet18) for each noise type. Each color indicates a noise level. Numbers below the noise names denote the absolute recognition accuracy at time-step 0 (feedforward baseline).}
    \label{supfig:presnet_shared_accuracy}
\end{figure}

\begin{figure}[H]
    \centering
    \includegraphics[width=\textwidth]{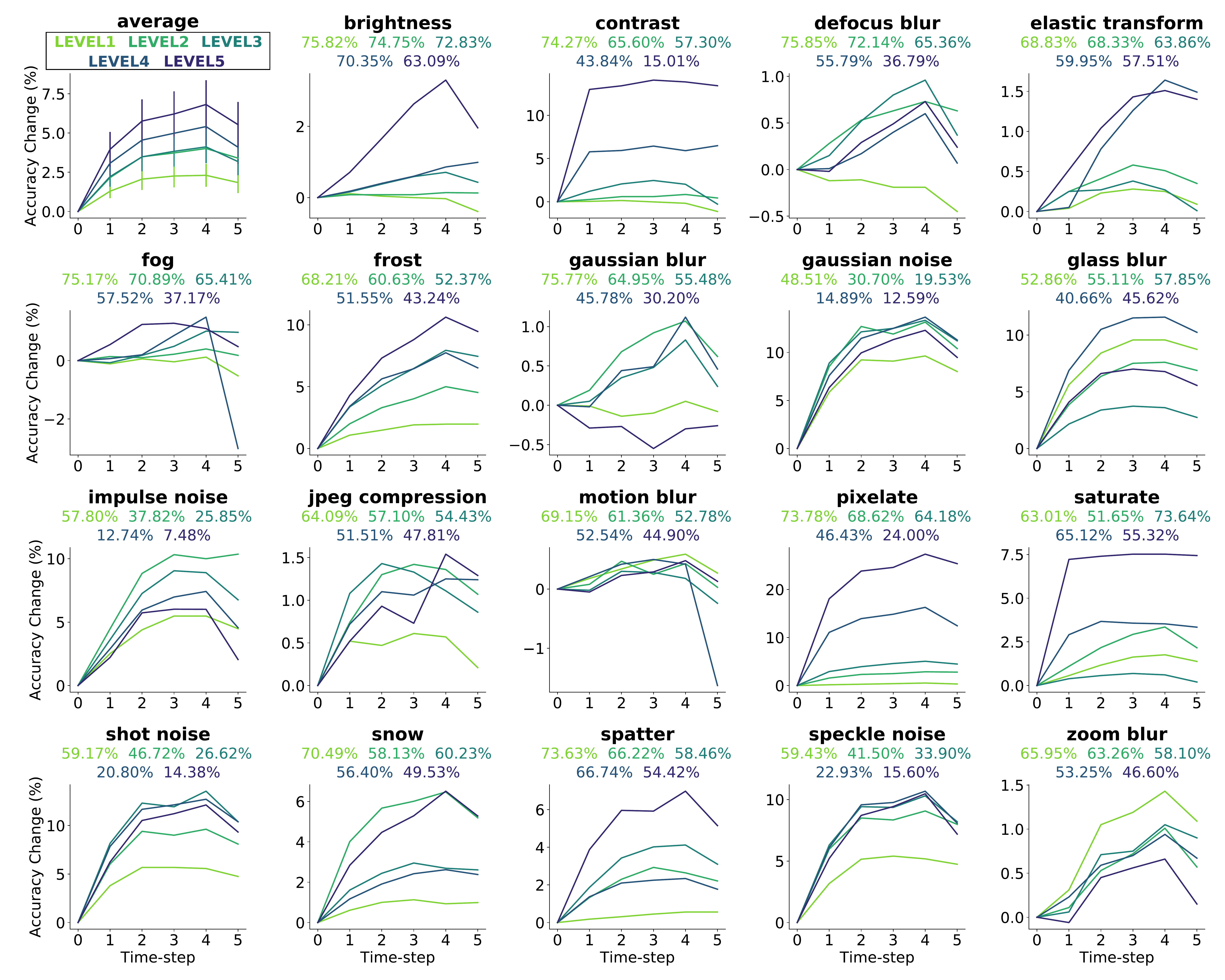}
    \caption{Recognition accuracy of PResNet18 with separate hyper-parameters per PCoder on each of the CIFAR100-C noise types and levels. Each plot shows the change in accuracy with respect to the feedforward baseline (i.e. ResNet18) for each noise type. Each color indicates a noise level. Numbers below the noise names denote the absolute recognition accuracy at time-step 0 (feedforward baseline).}
    \label{supfig:presnet_separate_accuracy}
\end{figure}

\begin{figure}[H]
    \centering
    \includegraphics[width=0.35\textwidth]{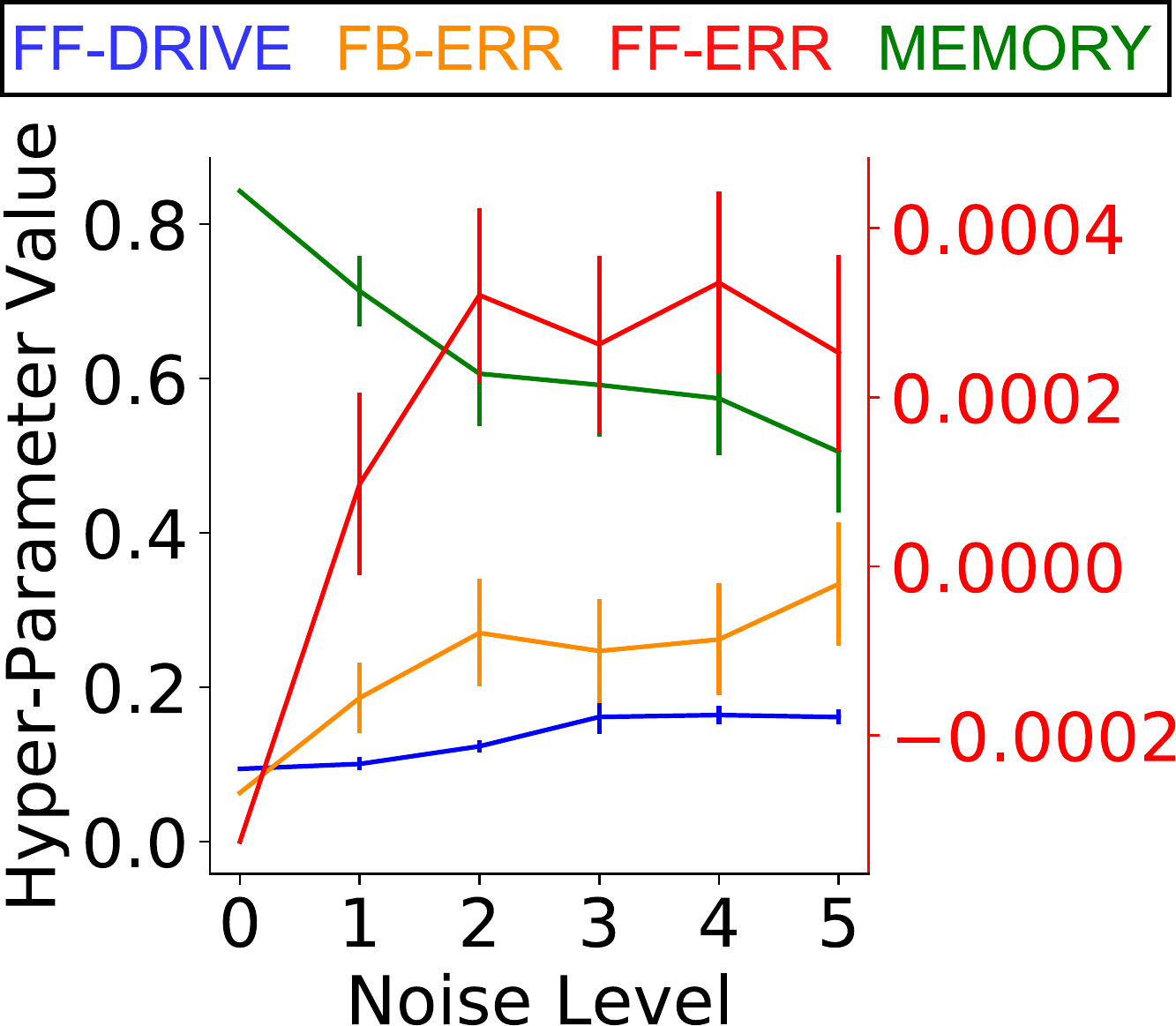}
    \caption{Absolute values of hyper-parameters of PResNet18 when they are shared among PCoders. Each line shows the average value of a hyper-parameter across all 19 CIFAR100-C noise types. Error bars indicate standard error of the mean. The value feedforward error hyper-parameter is plotted with a second y-axis (red). Noise level 0 denotes clean images.}
    \label{supfig:presnet_shared_hyperparameters}
\end{figure}

\begin{figure}[H]
    \centering
    \includegraphics[width=\textwidth]{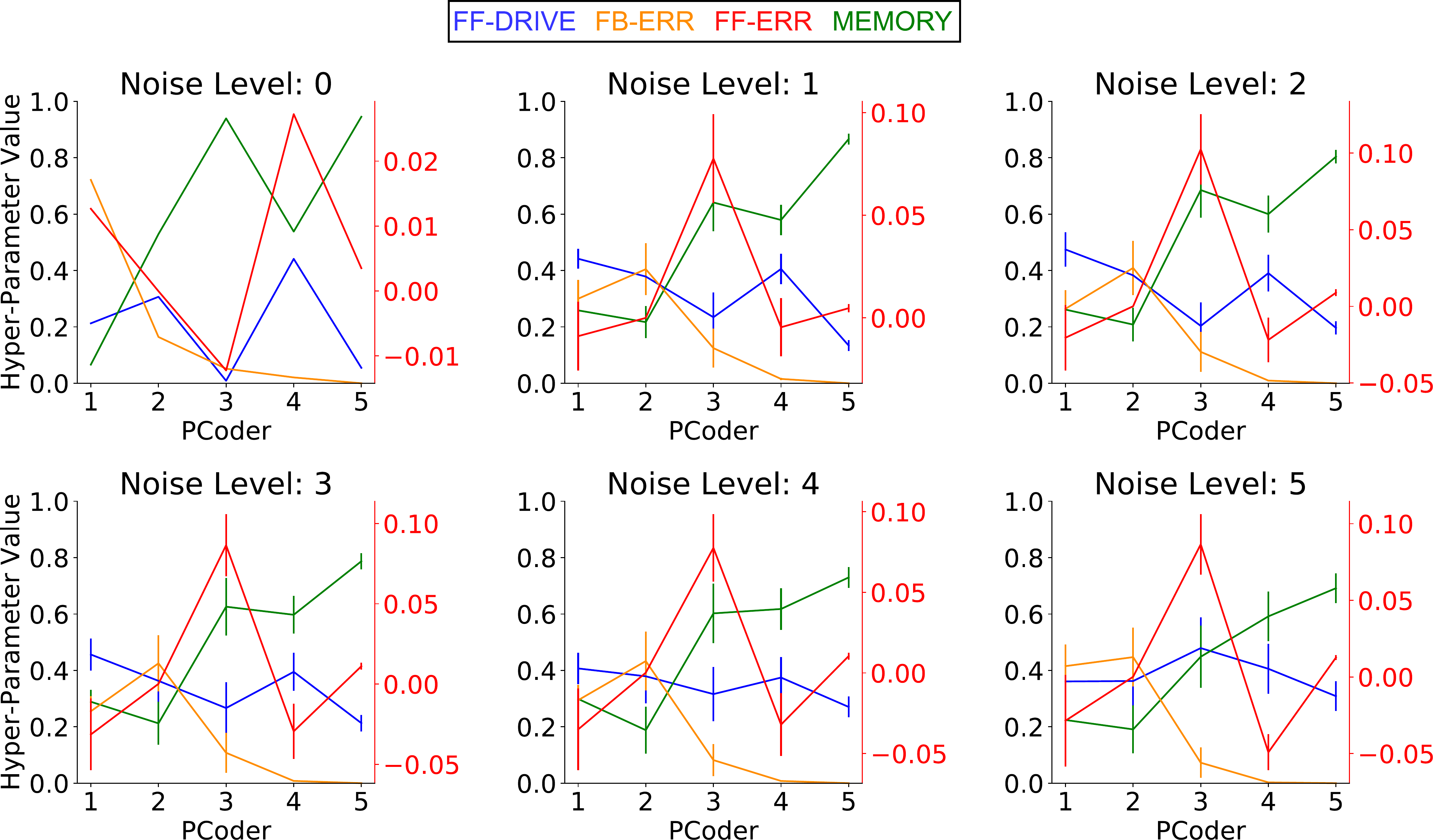}
    \caption{Absolute values of hyper-parameters of PResNet18 when each PCoder uses separate ones. Each plot shows the results of training hyper-parameters as a function of PCoders for a particular noise level. Each line shows the average value of a hyper-parameter across all 19 CIFAR100-C noise types. Error bars indicate standard error of the mean. The value feedforward error hyper-parameter is plotted with a second y-axis (red). Noise level 0 denotes clean images.}
    \label{supfig:presnet_separate_hyperparameters}
\end{figure}

\begin{figure}[H]
    \centering
    \includegraphics[width=\textwidth]{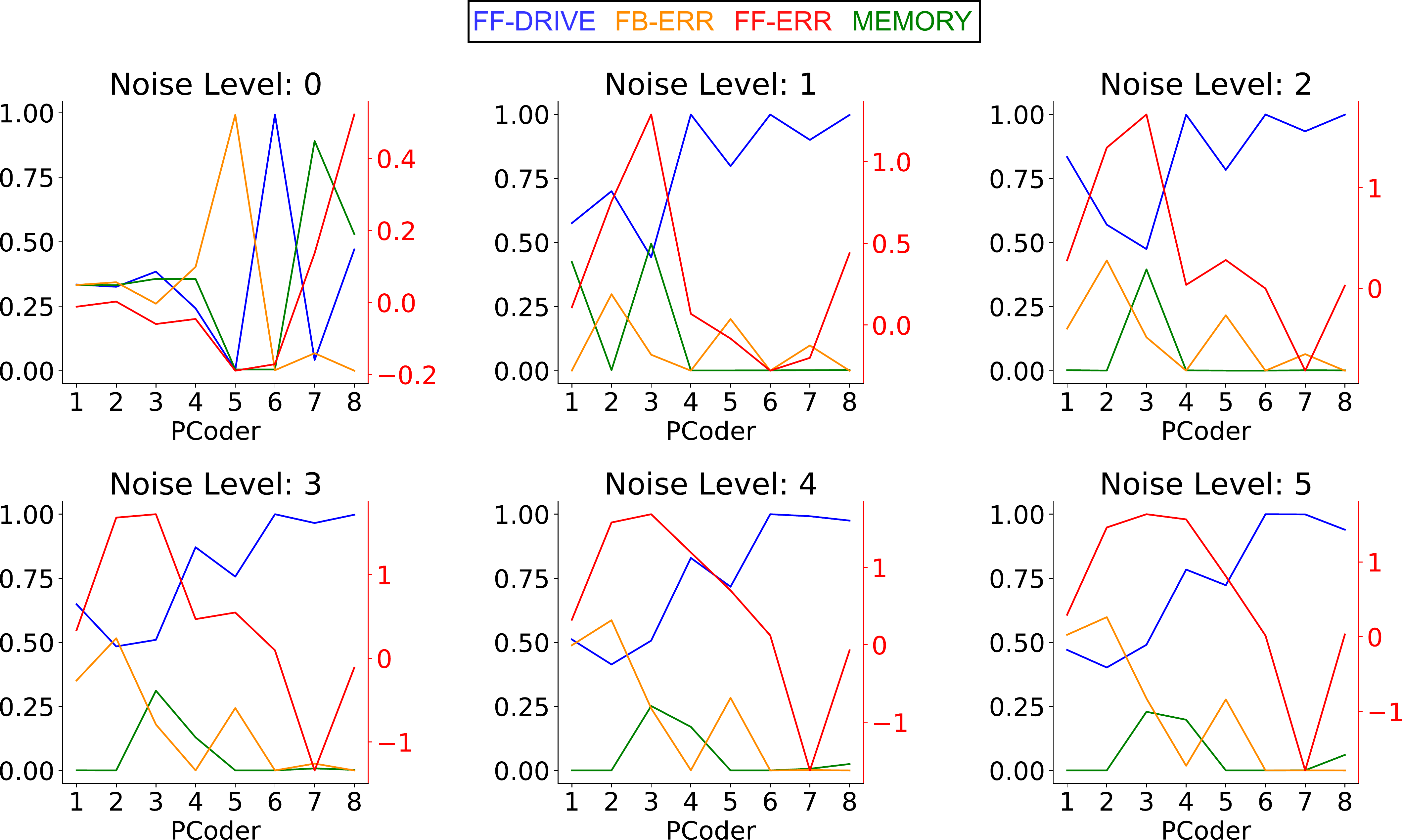}
    \caption{Absolute values of hyper-parameters of PEffNetB0 when each PCoder uses separate ones. Each plot shows the absolute value of a hyper-parameter as a function of PCoders for a particular level of Gaussian noise. The value feedforward error hyper-parameter is plotted with a second y-axis (red). Noise level 0 denotes clean images.}
    \label{supfig:peffnetb0_separate_hyperparameters_gaussian}
\end{figure}

\begin{figure}[H]
    \centering
    \includegraphics[width=\textwidth]{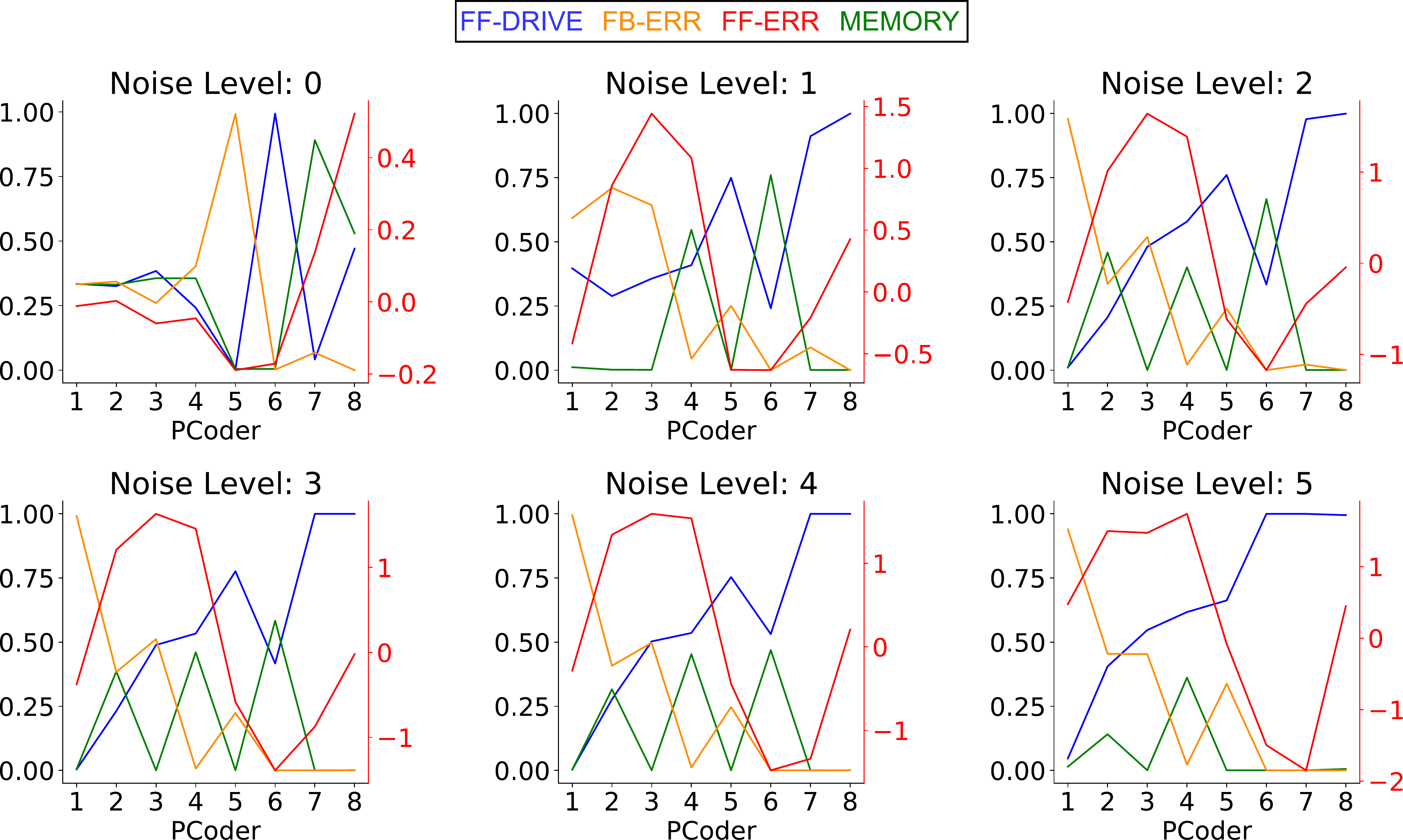}
    \caption{Absolute values of hyper-parameters of PEffNetB0 when each PCoder uses separate ones. Each plot shows the absolute value of a hyper-parameter as a function of PCoders for a particular level of Salt\&Pepper noise. The value feedforward error hyper-parameter is plotted with a second y-axis (red). Noise level 0 denotes clean images.}
    \label{supfig:peffnetb0_separate_hyperparameters_saltpepper}
\end{figure}

\end{document}